\documentclass[sigconf]{acmart}
\AtBeginDocument{%
  \providecommand\BibTeX{{%
    \normalfont B\kern-0.5em{\scshape i\kern-0.25em b}\kern-0.8em\TeX}}}

\begin{document}

\title{Automatic Seizure Detection Using the Pulse Transit Time}


\author{Eric Fiege}
\email{eric.fiege@isst.fraunhofer.de}
\orcid{1234-5678-9012}
\author{Salima Houta}
\email{salima.houta@isst.fraunhofer.de}
\orcid{1234-5678-9012}
\author{Pinar Bisgin}
\email{pinar.bisgin@isst.fraunhofer.de}
\orcid{1234-5678-9012}
\affiliation{
 \institution{Fraunhofer Institute for Software and Systems Engineering}
 \streetaddress{Emil-Figge-Str. 91}
 \city{Dortmund}
 \country{Germany}
}

\author{Rainer Surges}
\email{rainer.surges@ukbonn.de}
\affiliation{
 \institution{Department of Epileptology University of Bonn Medical Center}
 \city{Bonn}
 \country{Germany}}

\author{Falk Howar}
\email{falk.howar@tu-dortmund.de}
\affiliation{
 \institution{TU Dortmund University}
 \city{Dortmund}
 \country{Germany}}

\renewcommand{\shortauthors}{Fiege, et al.}

\newcommand{\FScore}{\textit{$F_1$-Score}}
\newcommand{\RRintervals}{RR-intervals}
\newcommand{\PTT}{PTT}
\newcommand{\PTTAlgorithm}{PTT Algorithm}
\newcommand{\ReactiveAlgorithm}{Reactive Algorithm}
\newcommand{\ECGPPGentries}{ECG and PPG entries}
\newcommand{\PPGentry}{PPG entry}
\newcommand{\ECGentry}{ECG entry}
\newcommand{\ECGPPGSignals}{ECG and PPG signals}
\newcommand{\PTTentry}{PTT entry}
\newcommand{\ROCCurves}{ROC curves}
\newcommand{\ClockDrift}{clock drift}

\newcommand{\ResultReactiveAlgSampleLength}{$8$}
\newcommand{\ResultReactiveAlgSampleOffset}{$30$}
\newcommand{\ResultReactiveAlgWindowCount}{zwei}
\newcommand{\ResultReactiveAlgPositivSamples}{$114$}
\newcommand{\ResultReactiveAlgNegativSamples}{$1886$}
\newcommand{\ResultReactiveAlgTime}{$266.67$}
\newcommand{\ResultReactiveAlgTrees}{$1600$}
\newcommand{\ResultReactiveAlgDepth}{$20$}
\newcommand{\ResultReactiveAlgSplit}{$10$}
\newcommand{\ResultReactiveAlgSensitivity}{$42.12 \pm 9.08 \percent$}
\newcommand{\ResultReactiveAlgSpecifity}{$\SI{99.69 \pm 0.34}{\percent}$}
\newcommand{\ResultReactiveAlgPrecision}{$\SI{87.56 \pm 13.41}{\percent}$}
\newcommand{\ResultReactiveAlgFScore}{$0.56 \pm 0.05$}
\newcommand{\ResultReactiveAlgAUC}{$\SI{0.83 \pm 0.03}{}$}

\newcommand\note[1]{\textcolor{green}{#1}}
\newcommand\fix[1]{\textcolor{orange}{#1}}
\newcommand\delete[1]{\textcolor{red}{#1}}
\begin{abstract}
Documentation of epileptic seizures plays an essential role in planning medical therapy. Solutions for automated epileptic seizure detection can help improve the current problem of incomplete and erroneous manual documentation of epileptic seizures. In recent years, a number of wearable sensors have been tested for this purpose. However, detecting seizures with subtle symptoms remains difficult and current solutions tend to have a high false alarm rate. Seizures can also affect the patient's arterial blood pressure, which has not yet been studied for detection with sensors. The pulse transit time (PTT) provides a noninvasive estimate of arterial blood pressure. It can be obtained by using to two sensors, which are measuring the time differences between arrivals of the pulse waves. Due to separated time chips a clock drift emerges, which is strongly influencing the PTT. In this work, we present an algorithm which responds to alterations in the PTT, considering the clock drift and enabling the noninvasive monitoring of blood pressure alterations using separated sensors. Furthermore we investigated whether seizures can be detected using the PTT. Our results indicate that using the algorithm, it is possible to detect seizures with a Random Forest. Using the PTT along with other signals in a multimodal approach, the detection of seizures with subtle symptoms could thereby be improved.

\end{abstract}






\maketitle

\section{Introduction}
%
%
%
%

Epilepsy is one of the most common neurological disorders worldwide and limits the autonomy of those affected by recurrent and unpredictable epileptic seizures. If epileptic seizures are associated with impaired consciousness and loss of control over bodily functions, the consequences can be life-threatening (e.g., failure of the respiratory center, accidents, suffocation). Taking appropriate safety precautions in time helps to avoid serious consequences or even sudden unexpected death in epilepsy (SUDEP). By using technological solutions to improve monitoring (e.g., video cameras, pulse oximeters) or accommodating relatives, SUDEP incidence in an epilepsy center shows a decreasing trend between $ 1981 $ and $ 2016 $. Accurate seizure recording also supports individualized drug therapy planning. Documentation is currently done in paper or web-based calendars (e.g., EPI-Vista®) \cite {epivista}. However, studies show that about $50\%$ of seizures are not documented and about two-thirds of patients provide incorrect information \cite{blum1996patient, hoppe2007epilepsy}. 
The main reasons for incorrect seizure documentation include impaired perception of one's own seizures, error-prone seizure documentation by relatives or caregivers \cite{akman2009seizure, Nijsen2005}. 
Sensor-based systems that can accurately detect epileptic seizures are therefore of great importance and are being increasingly explored. 

Seizures can affect brain regions that contribute to regulation of autonomic body functions such as heart rate, respiratory rate, sweating, and blood pressure \cite{Moseley2013, bpFocalEpilepsy, Nass2019}. 
While motor phenomena, heart rate changes, and sympathetic skin response have been used to detect seizures \cite{Vandecasteele2017,Poh2012} with wearable sensors, the detection of seizures with blood pressure (BP) has not been investigated.
Current sensors do not use this information because noninvasive measurement of BP is challenging. The PTT is directly related to the BP and can be determined noninvasively \cite{Hennig2013}. It describes the time required for a pulse wave to arrive in a peripheral part of the body. 
In this work we investigate if seizures could automatically be detected by using the \PTT.

\section{Related Work}
The gold standard for seizure detection is a combination of video and electroencephalography (video-EEG). An EEG monitor can reliably detect seizures using analysis of brain signals \cite{Nigam2004, Tzallas2009a, Jaiswal2018}. Intracranial EEG sensors provide the most reliable signals but are very invasive, not suitable for all forms of epilepsy, and are less well accepted by patients \cite{Hoppe2015}. Because existing EEG measurement solutions are not practical for use in the patient's home, research on mobile solutions using other biomarkers has been ongoing for several years. Depending on the seizure type, each of which is associated with different accompanying symptoms, this research usually focuses on detecting a group of similar seizure types. Different research works provide a literature review of seizure detection devices and their effectiveness for different seizure types \cite{Ulate-Campos2016, Kurada2019}. Seizures involving the muscles can be well detected with acceleration data. \cite{Beniczky2013, Embrace2016} Many present sensors are used for the detection of motor seizures, using acceleration data. However, there are also nonmotor seizures that cannot be detected with acceleration data. Seizures also affect the autonomic nervous system and the processes controlled by it. With EDA, HR or BP, therefore, there are other signals, that can be utilized to detect seizures. The difficulty is that changes in these signals occur frequently, without a seizure, in the everyday life of the patient, leading to a high false-alert rate. Under the use of multiple signals the false alarms may be decreased.

The effects on the cardiovascular system have been investigated in many studies. Some seizures may affect the patient's HR. In most cases, they result in tachycardia, rarely in bradycardia. The rise in HR may take a few seconds before the seizure is visible in the EEG. The time period varies depending on the type of present epilepsy. In addition, one in five patients is known to have irregularities in the EEG \cite{Leutmezer2003, Opherk2002}.
Studies have shown that seizures can affect also the BP \cite{bpFocalEpilepsy, Nass2019}. However, current sensors do not use this information because noninvasive measurement of BP is still a major challenge. The PTT is directly related to the BP and
can be determined non invasively \cite {Hennig2013}. One approach to determine PTT is to measure the time between cardiac contraction with an ECG and arrival in the peripheral part of the body with an PPG sensor. In this work, the PPG signal was recorded from a sensor placed in the patient's inner ear. The PTT is calculated by subtracting the start times of RR intervals from the ECG and PPG signals. 

\section{Methods}


We created a data processing pipeline to structure the process of data refinement and to grant insight for validation. The processing steps can be categorized into three major categories. At first, we loaded and prepared \RRintervals{} of both ECG and PPG signals. A matching combination of those signals will be called a "case". Next, we calculated and further improved the PTT for all cases. Finally, we generated the data set, trained a Random Forest and calculated quality measures. The following sections describe the clinical study and the data basis. Furthermore we describe the mentioned major categories in detail. Finally, we present the methods used to validate the performance of the trained models.

\subsection{Clinical Study}
Clinical trials are being conducted at participating specialty clinics for derivation of biosignal patterns and algorithm development and validation. A total of $200$ patients were recruited to test the in-ear sensor. Video EEG, ECG, and in-ear sensor data (PPG, HR, temperature, acceleration data) were collected from these patients over an average period of four days. Based on the video EEG, physicians recorded seizures that occurred (time period, type of seizure). Through July $2019$, $552$ seizures have been recorded. The collected data form the basis for the present work.

\subsection{RR interval Preprocessing}
\begin{figure}
	\includegraphics[width=0.5\textwidth]{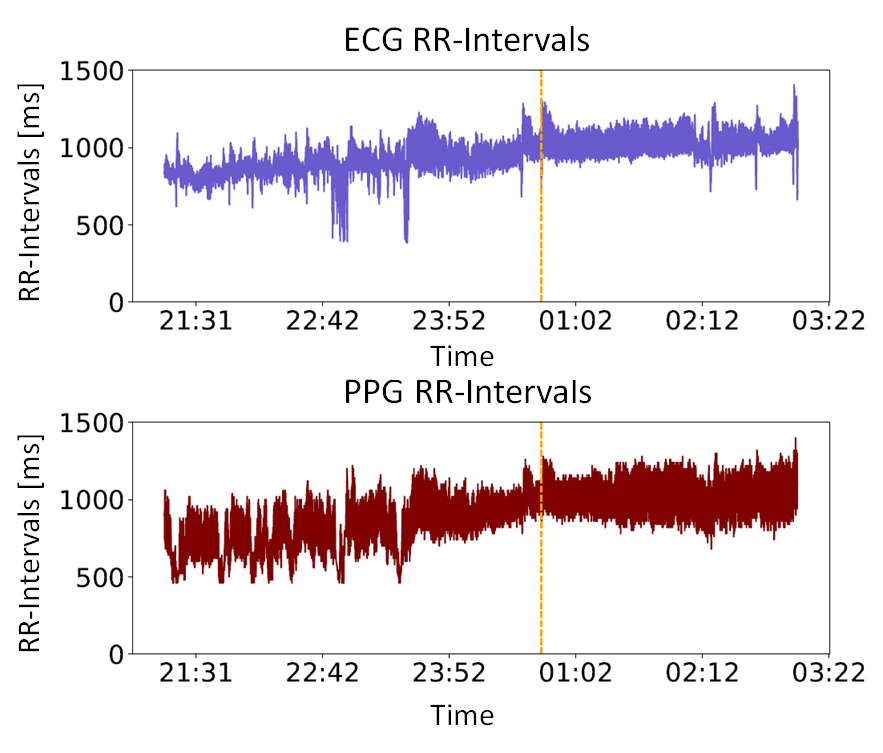}
	\centering
	\caption{Result of RR interval preprocessing. The outliers have been removed and only parallel recorded data has been kept. A seizure is marked by the dashed orange line.}
	\label{fig:improvedrecordings}
\end{figure}

To find the parallel recorded \ECGPPGSignals, first we loaded all metadata files into a database and queried the parallel recorded ECG and PPG filenames from the database. Then we loaded the raw data from the files.
The raw data contains a lot of outliers so, as a first refinement step, we removed data points, if their difference to the empirical mean deviates more than twice of the standard deviation on the entire signal.
The parallel recorded \ECGPPGSignals{} also do not start and stop at the same time. But for the calculation of the PTT it is mandatory to have both signals. Therefore we clipped the parallel recorded signals to the overlapping data. Figure \ref{fig:improvedrecordings} shows the result of the RR interval preprocessing.

Some cases show a systematic error in the RR interval calculation, resulting in a great difference between the ECG and PPG interval lengths and are therefore removed. The condition for a removal is that the difference of their empirical mean values is greater than a constant $k$. In a real world application the model should not be evaluated if the \PTT{} cannot be calculated properly.

\subsection{Calculating the PTT}
The preprocessed RR intervals could be used to calculate the PTT. But there must be considered an effect occurring due to the fact that each recording device has had its own time chip. This leads to a temporal offset between the \ECGPPGSignals, increasing with the length of the recordings. This circumstance is called \ClockDrift{} and has a huge impact on the \PTT. As a result the calculation of the absolute \PTT{} with the collected data is not possible. Relative alterations on the other side can be reconstructed. To reconstruct the relative alterations the \ClockDrift{} has to be removed as good as possible. This however always effects the \PTT. It is therefore important to have a good balance between removing the \ClockDrift{} and keeping the alterations in the \PTT. If too much clock drift is removed also alterations in the \PTT{} are removed and seizures may no longer be detected. In contrast, if a too small amount of clock drift is removed, the changes in the \PTT{} are masked by the \ClockDrift. 

In this work we developed two algorithms. The first algorithm (\PTTAlgorithm) calculates the \PTT{} on small time intervals and will ignore the emerging \ClockDrift{} between both signals. The second algorithm (\ReactiveAlgorithm) uses the first algorithm and tries to remove the \ClockDrift{} on longer time intervals. The output of this algorithm is a time series that responds to alterations in the \PTT.

The \PTTAlgorithm{} searches for matching \ECGPPGentries. Each \PPGentry{} $(t_{PPG}, RR_{PPG})$, where $t_{PPG}$ is the occurrence time of the PPG RR interval and $RR_{PPG}$ is the length of the PPG RR interval, which follows directly on an \ECGentry{} $(t_{ECG}, RR_{ECG})$ will be used to calculate the $i$-th \PTT{} entry as follows:
\begin{equation}
PTT_i \coloneqq (t_{PTT}, PTT) = (t_{ECG}, t_{PPG} - t_{ECG}) \nonumber
\end{equation}
If the resulting \PTT{} is greater than $RR_{ECG}$ the resulting \PTTentry{} is discarded. With perfectly aligned data tracks and without \ClockDrift{} this algorithm would calculate the absolute \PTT. 

The idea behind the \ReactiveAlgorithm{} is to shift the start times of all PPG entries by an offset $o$, in such way that for a limited time interval the calculated \PTT{} values are located in the middle of the ECG RR intervals. For each \ECGentry{} the following next few minutes of data are selected as a window and the \PTT{} is calculated with the \PTTAlgorithm. For a new \PTT{} data point, the mean value of the calculated \PTT{} values on the window is used. The data point is only used, if $20\%$ of the expected ECG- and PPG intervals are available. After each calculation, the offset of the PPG time series is adjusted. This is accomplished by subtracting the new PTT data point value from the half of the mean of the ECG RR interval length. The mean value $ ecg\_mean\_rr\_interval $ of the ECG RR interval is based on the entire length of the time series, whereas the new PTT data point value $w_{PTT}$ is calculated on a window of a few minutes. The PPG-offset is than given by:
\begin{equation}
o \coloneqq \frac{ecg\_mean\_rr\_interval/2 - w_{PTT}}{c}. \nonumber
\end{equation}
The constant $c$ can be used to control how much \ClockDrift{} is removed. With an increasing $c$ the algorithm removes less \ClockDrift. All in all it is ensured that the PPG start times are as centered as possible between the ECG start times for all windows.

\begin{figure}
	\includegraphics[width=0.5\textwidth]{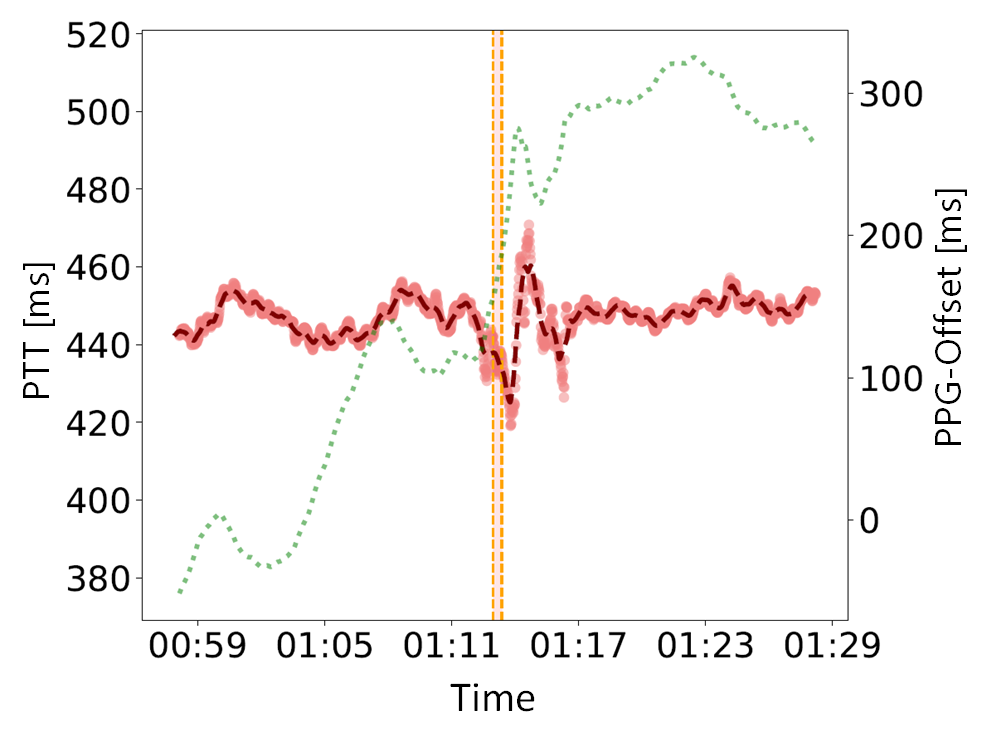}
	\centering
	\caption{PTT-alterations calculated by the \ReactiveAlgorithm{} for a CPS seizure. The green line shows the added PPG-offset during the calculation.\label{imgSampleCaseReactive}}
	\vspace{-0.5cm}
\end{figure}

\subsection{Feature Generation and Training of Models}
For those cases that contain a seizure, there was exactly one training pair generated. Cases without seizure where used to create as many non-overlapping training pairs as possible. We created the training pairs on a fixed time interval and divided this time interval into windows. The window count can be set to two or three windows. In addition, there may be added a time shift $o_s$ for cases which contain a seizure. This shift should ensure that a sudden change in the PTT is not located between two windows.

We generated the positive samples as follows. First we determined the temporal middle $t_m$ of the seizure by the start and end time of the seizure. After that, with $ t_c = t_m + o_s $, the shift is added. Finally, we selected the PTT around $ t_c $ with the interval length $l$. If the start or end time of the selected PTT is outside of the recorded PTT, we discarded the training pair.  

The input variables of the samples are based on statistical quantities derived from the PTT and are calculated for each window. The length of the samples inner windows depends on their number and is chosen in such way, that all of them have the same length. For each window we calculated the mean, minimum, maximum, variance, slope and the intercept of the regression line. Furthermore we used the differences of the mean values of all windows. All these values serve as input variables for the model. Finally we added the target variable to the training pair. 

With the resulting dataset a Random Forest has been trained and tested. We optimized the hyperparameters, including the minimum of samples required to split a node, the tree count and depth of trees. Possible parameters are drawn from an equally distributed set of numbers. We then trained the Random Forest with the drawn selection and tested with five-fold cross-validation. 

\subsection{Validation of the Performance}
In order to validate the performance of our detection model we use the method proposed by Beniczky et. al. \cite{Beniczky2018}. To estimate the accuracy and performance of seizure detection, Beniczky et al. define which outcome measures are to be considered. The parameters also refer to the handling and error-prone nature of the device. The focus of this work is the basic applicability of the \PTT{} for the detection and does not consider an explicit device. Consequently, we limited our evaluation to the following outcome measures:
\begin{itemize}
	\item Sensitivity number (and percentage) of all detected seizures/ number of all seizures recorded during the study
	\item False alarm rate: the number of false alarms per $24$ hours
	\item Detection latency: time from seizure onset to the detection time
\end{itemize}
In addition we use the Receiver-Operating-Characteristic (ROC) curve and the \FScore{} to evaluate the classifier performance \cite{zweig1993}. 

\section{Results}

The generated database contained $239$ parallel recorded \ECGPPGSignals. A seizure occurred in $149$ cases. 
The constant $k$, used for the removal of cases with improper calculated RR-intervals, has been set to $200$ milliseconds. In this way $34$ cases were excluded from the further data processing. We calculated the PTT using both \PTTAlgorithm{} and \ReactiveAlgorithm. Figure \ref{imgSampleCaseReactive} shows an example for the calculated \PTT-alterations using the \ReactiveAlgorithm{} during a CPS.

Using a window size of $8$ minutes, the generated data set contained $114$ positive and $1886$ negative samples. There were nine different seizure types included and the features were generated using two inner windows. The $o_s$ parameter has been set to $30$ seconds. On this dataset we trained and tested a Random Forest as a binary classifier with five-fold cross-validation. We used the parameters with the best \FScore{} for a final evaluation. The following hyperparameters are the results of the optimization:
\begin{itemize}
	\item Number of Trees: $1600$
	\item Maximum depth: $20$
	\item Minimum node split: $10$
\end{itemize}

\noindent Table \ref{tblAlgPerformance} shows the overall classification performance for both \PTTAlgorithm{} and \ReactiveAlgorithm. The \ReactiveAlgorithm{} yields better results than the \PTTAlgorithm.
\begin{table}[htbp]
	\caption{Measures of the Performance for both \PTTAlgorithm{} and \ReactiveAlgorithm{} using a Random Forest}\label{tblAlgPerformance}
	\centering%
	\begin{tabular}{lccc}
		\hline
		 & \textit{\PTTAlgorithm} & \textit{\ReactiveAlgorithm} \\
		\hline
		$Sensitivity$ & $16.39\% \pm 13.29\%$ & $42.12\% \pm 9.08\%$ \\
		$Specificity$ & $98.84\% \pm 0.01\%$ & $99.69\% \pm 0.34\%$ \\
		$Precision$ & $28.97\% \pm 10.83\%$ & $84.56\% \pm 13.41\%$ \\
		\FScore & $0.20 \pm 0.13$ & $0.56 \pm 0.05$ \\
		\hline
	\end{tabular}
\end{table}

\begin{figure}
	\includegraphics[width=0.45\textwidth]{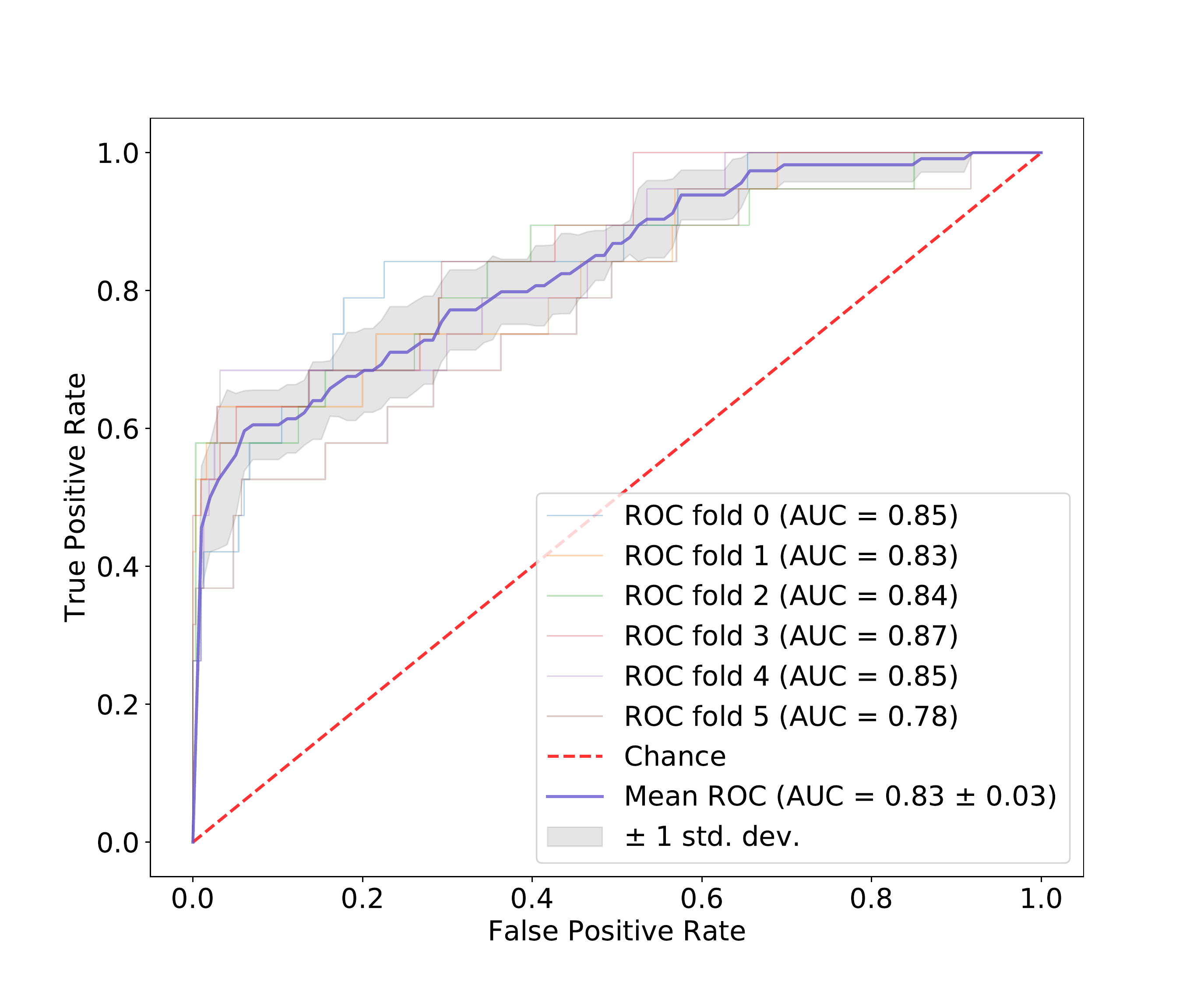}
	\caption[]{ROC curve under the usage of the \ReactiveAlgorithm{} and a Random Forest. The Random Forest is in $40$\% of the seizures confident that it is a seizure. The rest of the seizures is more difficult to distinguish from the negative training pairs. \label{imgModelReactiveAlgROC}}
\end{figure}
The ROC curve can be seen in Figure \ref{imgModelReactiveAlgROC} and starts with a strong increase until over $40$ \% of the seizures have been classified correctly. Thus, the model is in $40$\% of the seizures confident that it is a seizure. The rest of the seizures is more difficult to distinguish from the negative training pairs. The further course of the ROC curve indicates a random process. 

Table \ref{tblClasses} shows the detection performance for specific seizure types. The best model, based on the \ReactiveAlgorithm, identified $37.31$ \% of CPS, $21.43$ \% of SPS, and $20.00$ \% of absence seizures. This shows that seizures can be detected with the PTT. The classification results however depend on the seizures types.
The seizures involving the muscles ($10$ GTCS, $4$ tonic and $1$ atonic) were all correctly detected. An exception to this are the myoclonic seizures, which could not be detected. 
\begin{table}[htbp]
	\caption{Detection Performance for Seizures Types using the \ReactiveAlgorithm{} and a Random Forest. \label{tblClasses}}
	\centering%
	\begin{tabular}{lccc}
		\hline
		\textit{Type} & \textit{Correct} & \textit{Total} & \textit{Rate}\\
		\hline
		$GTCS$ & $10$ & $10$ & $100.00$ \% \\
		$tonic$ & $4$ & $4$ & $100.00$\%\\
		$atonic$ & $1$ & $1$ & $100.00$\%\\
		$CPS$ & $25$ & $67$ & $37.31$\%\\
		$absence$ & $1$ & $5$ & $20.00$\%\\
		$aura$ & $4$ & $23$ & $17.39$\%\\
		$myoclonic$ & $0$ & $4$ & $0.00$\%\\
		\hline
	\end{tabular}
\end{table}
The high detection rate for motor seizures could be caused by a deterioration in data quality. This is supported by the fact that the model uses the variance of the training pairs to separate the classes. However, the increased variance could also be caused by the valley output by the reactive algorithm when the PTT drops. It should also be noted that the \ReactiveAlgorithm{} can lead to increased variance if the heart rate is increased. Accordingly, the model could not only predict the seizures from the change in PTT, but also an increased heart rate. The myoclonic seizures, even though they have not been recognized by the model, could also affect blood pressure. They could lead to minor alterations in the \PTT{} and not be detected cause the removal of the clock drift filtered those alterations too.

With a number of six false-positive training pairs, out of $266.67$ hours of data, the false alarm rate is $0.54$ alarms per day. The true false alarm rate however may be higher, because sections with a great difference between ECG and PPG interval lengths were excluded. Due to the usage of $8$ minute windows and $o_s = 30s$ the detection latency is $210$ seconds.

\section{Conclusion and Future Work}
The developed PTT algorithm takes the clock drift into account and responds to alterations in the PTT. The degree of respondence to alterations and removal of the clock-drift can be adjusted.  
The results show that seizures can be detected by using the \PTT. Using the algorithm, which is considering the clock drift, the sensitivity increased from $16.39\% \pm 13.29\%$ to $42.12\% \pm 9.08\%$ compared to the simple PTT algorithm. The recognition rate depends on the type of seizure. 
The best model was trained under use of a Random Forest and detected almost all motor seizures. Furthermore, it detected $37.31$ \% of the CPS and $21.43$ \% of the SPS. The \FScore{} is \ResultReactiveAlgFScore.
The Random Forest separates seizures mainly through the variance and the minimum, for samples of $8$ minutes. The high detection rate with motor seizures can have two reasons. On the one hand, the blood pressure in this kind of seizures could fall more sharply. The separation over the variance could then be explained by the increased variance caused by the valley. On the other hand, the increased variance can also be explained by measurement errors that arise from the movements during the seizure. The Random Forest could detect seizures, not only by the \PTT, but also by a worse data quality. 

In future work, besides the use of a Random Forest, other methods can also be used for model prediction. Furthermore the training of models for specific seizures types could lead to a higher sensitivity. Features generated using the PTT should be evaluated in a multimodal approach and may reduce the false alarm rate. In addition to seizure detection, also other diseases can be focused on using the same sensor technology and taking alterations of blood pressure into account.


\bibliographystyle{ACM-Reference-Format}
\bibliography{sample-base}


\end{document}